\documentclass[preprint,12pt]{elsarticle}
\usepackage{hyperref}

\usepackage{amssymb}
\usepackage{amsmath}

\usepackage[utf8]{inputenc}     
\usepackage[T1]{fontenc}        
\usepackage{lmodern}            
\usepackage{geometry}           
\usepackage{graphicx}           
\usepackage{hyperref}           
\usepackage{amsmath}            
\usepackage{amssymb}            
\usepackage{cleveref}           
\usepackage{float}              
\usepackage{booktabs}           
\usepackage{array}              
\usepackage{caption}            
\usepackage{subcaption}         
\usepackage{color}              
\usepackage{url}                
\usepackage{enumitem}           
\usepackage{lineno}             
\usepackage{appendix}           
\usepackage{multirow}           
\usepackage{fancyhdr}           
\usepackage{multicol}           

\usepackage{pgfplots}           
\pgfplotsset{compat=1.18}       
\usepackage{tikz}               
\usetikzlibrary{positioning, shapes, arrows, calc, intersections, patterns, matrix}
\usepackage{pgfplotstable}      

\usepackage[normalem]{ulem}
\usepackage{enumitem}
\usepackage{wrapfig}
\usepackage{caption}

\usepackage{cuted}

\usepackage{algorithm}
\usepackage{algpseudocode} 

\usepackage{natbib}           

\setcitestyle{authoryear,round}

\usepackage{xcolor}             
\usepackage{listings}           
\lstdefinelanguage{json}{
    basicstyle=\ttfamily\footnotesize,
    showstringspaces=false,
    breaklines=true,
    frame=single,
    backgroundcolor=\color{gray!10},
    string=[s]{"}{"},
    stringstyle=\color{red},
    morestring=[b]',
    identifierstyle=\color{blue},
    comment=[l]{//},
    commentstyle=\color{green},
    keywordstyle=\color{black}
}

\usepackage{siunitx}
\journal{Advances in Space Research}

\begin{document}

\begin{frontmatter}



\title{Large Language Models as Autonomous Spacecraft Operators in Kerbal Space Program}


\author[label1]{Alejandro Carrasco}
\affiliation[label1]{
    organization={Massachusetts Institute of Technology, Department of Aeronautics and Astronautics},
    addressline={77 Massachusetts Avenue},
    city={Cambridge},
    state={MA},
    postcode={02139},
    country={USA}
}

\author[label1,label2]{Victor Rodriguez-Fernandez}
\author[label1]{Richard Linares}

\affiliation[label2]{
    organization={Universidad Politécnica de Madrid},
    addressline={Escuela Técnica Superior de Ingeniería de Sistemas Informáticos},
    city={Madrid},
    postcode={28031},
    state={Madrid},
    country={Spain}
}

\begin{abstract}
Recent trends are emerging in the use of Large Language Models (LLMs) as autonomous agents that take actions based on the content of the user text prompts. We intend to apply these concepts to the field of Control in space, enabling LLMs to play a significant role in the decision-making process for autonomous satellite operations. As a first step towards this goal, we have developed a pure LLM-based solution for the Kerbal Space Program Differential Games (KSPDG) challenge, a public software design competition where participants create autonomous agents for maneuvering satellites involved in non-cooperative space operations, running on the KSP game engine. Our approach leverages prompt engineering, few-shot prompting, and fine-tuning techniques to create an effective LLM-based agent that ranked 2nd in the competition. To the best of our knowledge, this work pioneers the integration of LLM agents into space research. The project comprises several open repositories to facilitate replication and further research. The codebase is accessible on \href{https://github.com/ARCLab-MIT/kspdg}{GitHub}, while the trained models and datasets are available on \href{https://huggingface.co/OhhTuRnz}{Hugging Face}. Additionally, experiment tracking and detailed results can be reviewed on \href{https://wandb.ai/carrusk/huggingface}{Weights \& Biases}.
\end{abstract}

\begin{graphicalabstract}
\end{graphicalabstract}
\begin{keyword}
Large Language Models \sep Autonomous Agents \sep Kerbal Space Program \sep Prompt Engineering \sep Fine-tuning \sep Spacecraft Control
\end{keyword}

\end{frontmatter}

\section{Introduction}
Large Language Models (LLMs) are, without a doubt, the last major breakthrough in the evolution of artificial intelligence systems. Since the release of ChatGPT \cite{chatgpt} at the end of 2022, we have seen a plethora of applications and use cases emerge across various industries. From generating human-like text to aiding in code completion, LLMs have significantly impacted the way we interact with technology and the possibilities of what AI can achieve.

In recent times, LLMs have progressed from text-based applications to \textit{language agents} capable of taking context-aware actions, as demonstrated by Anthropic's computer use agent\footnote{\url{https://www.anthropic.com/news/3-5-models-and-computer-use}}. By leveraging contextual information, LLMs can autonomously perform tasks which usually correspond to software-based systems. Expanding on this, these so-called agents extend beyond software applications, as researchers are applying this concept to physical systems such as LLM-driven robots \cite{wang2023survey} and self-driving car motion planners \cite{mao2023gpt}.


Historically, spacecraft Guidance, Navigation, and Control (GNC) systems have relied on classical control methods like Proportional-Integral-Derivative (PID) control, Linear Quadratic Regulators (LQR), and more advanced techniques such as Model Predictive Control (MPC) and robust control. While these methods have ensured reliable spacecraft operations, they often require extensive tuning and precise dynamic modeling. AI and machine learning, particularly Reinforcement Learning (RL), have emerged as promising alternatives, offering more adaptive solutions that optimize control policies through environmental interaction, with notable examples including agents trained for tasks such as sensor-tasking \cite{siew2022space} and planetary landing \cite{gaudet2020deep}.


This work focuses on space applications, particularly the development of autonomous agents for the guidance and control of spacecraft. However, unlike other fields of AI research, the space domain suffers from a lack of publicly available simulation environments, which are essential for training AI agents in complex scenarios and benchmarking autonomous control methods. To address the lack of simulation environments in space applications, Allen et al. introduced \textit{SpaceGym} \cite{10115968}, which includes the Kerbal Space Program Differential Games suite (KSPDG)—the primary focus of this work.


However, KSPDG is unsuitable for RL training due to technical and design constraints. The KSP engine lacks capabilities for parallel, accelerated, and headless operations necessary for faster-than-real-time RL training. Moreover, KSPDG's creators intentionally designed it as an evaluation framework, a “true test set” environment that minimizes overfitting and prioritizes unbiased testing of AI agent capabilities, diverging from RL’s iterative training paradigm. In contrast with traditional RL, which often struggles with sample inefficiency and the need for explicit reward functions, LLMs offer a promising alternative. Studies have shown that advanced LLMs, like GPT-4, can outperform state-of-the-art RL algorithms in complex games simply by analyzing academic texts and reasoning \cite{wu2023spring}, achieving sophisticated trajectories and strong zero-shot performance.

\begin{figure}[htbp]
    \centering
    \includegraphics[width=0.9\linewidth]{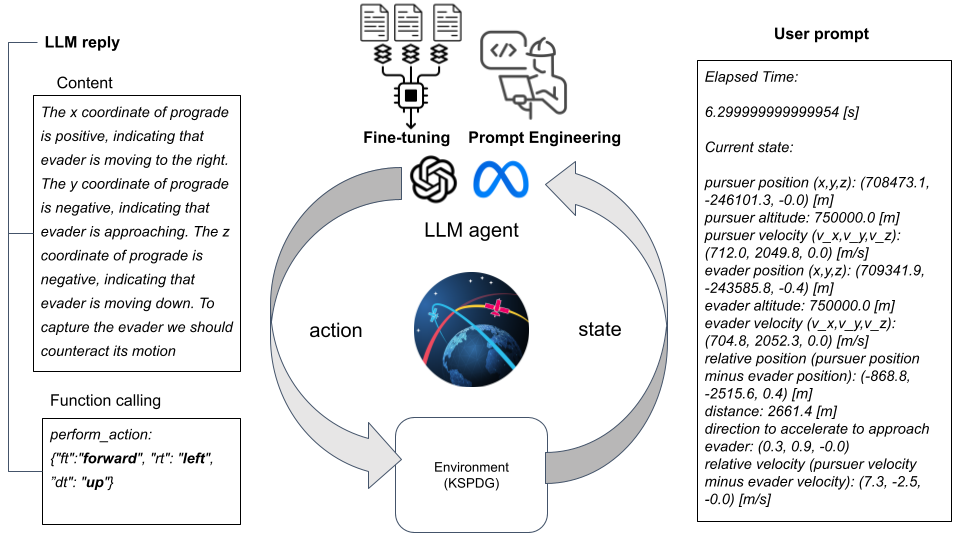}
    \caption{Overview of both LLM strategies based on the KSPDG Challenge agent implementation.}
    \label{fig:ksp_overview}
\end{figure}

To overcome the limitations of RL in creating autonomous agents for environments such as KSDPG, as well as for other space operations where numerous simulated data cannot be provided, we propose to adapt the current trend of LLM-based agents to develop an ``intelligent'' operator that controls a spacecraft based on the real-time telemetry of the environment, relying solely on natural language for both input and output. As depicted in both figures in \cref{fig:ksp_overview}, we design the classic RL loop by interfacing the simulation environment (KSDPG) with a LLM, transforming the real-time observations (or state) of the mission as textual user prompts that are fed to the model. The LLM then processes the prompt and replies with an action that will be plugged in KSDPG to control the spacecraft. In this work, we leverage GPT-3.5 and LLaMA models to develop an intelligent agent capable of controlling a spacecraft in real time. We explored various strategies to present the mission environment observations as textual inputs to the models and refined their performance using data collected from human gameplay and expert-mimicking bots. In both editions of KSPDG, our LLM centered agents were the only non-traditional optimization method to reach the podium, consistently advancing to the finals and securing second place consecutively \footnote{\href{https://www.ll.mit.edu/conferences-events/2024/01/kerbal-space-program-differential-game-challenge}{Kerbal Space Program Differential Game Challenge}}.

When deciding on an LLM approach, there are three primary strategies: creating a new model, prompt-engineering to leverage the capabilities of an existing model, or fine-tuning a model to optimize performance. Given the technical and resource constraints of this work, creating a new LLM is not feasible, though such a requirement is rare in practice. Instead, this study focuses on leveraging existing models through prompt-engineering and fine-tuning, which serve as viable and efficient alternatives. Fine-tuning, in particular, offers the opportunity to address specific limitations of base models, such as response latency or consistency, enabling the refinement of models to ensure they perform optimally for targeted tasks.

The LLM landscape has evolved rapidly over the past two years, with significant advancements and frequent updates leading to the emergence of models that rival the once-dominant ChatGPT. Notable examples include Claude \cite{anthropic2023claude}, Gemini \cite{deepmind2023gemini}, and open-source models like Mistral \cite{mistral2023mistral7b}, DeepSeek \cite{deepseek2024} and LLaMA \cite{touvron2023llama}. This study uses GPT for its ease of use and LLaMA for its community support and open-source flexibility.. As illustrated in \cref{fig:ksp_overview}, two strategies are compared: one involving prompt-engineering, and another using fine-tuned models.

\section{Large Language Models (LLMs)}


The 21st century has seen an explosion of interest and progress in AI (Computer Vision, Sentiment Analysis, Natural Language Processing). However, the development of large language models (LLMs) has been unprecedented. OpenAI's GPT-3 \cite{brown2020language}, released in 2020, demonstrated remarkable capabilities in generating human-like text and performing complex tasks based on textual input.

LLMs like GPT-4 \cite{openai2023gpt4}, Claude \cite{anthropic2023claude}, Gemini \cite{deepmind2023gemini}, Mistral \cite{mistral2023mistral7b}, and LLaMA \cite{touvron2023llama} have further pushed the boundaries of what AI can achieve. These models leverage vast amounts of data and computational power, resulting in great performance across a range of applications, from writing essays to generating code. The field continues to evolve rapidly, with ongoing research and development aimed at enhancing the capabilities and applications of AI \cite{arslanian2019understanding}.

Large language models are designed to understand and generate human language by processing massive datasets containing diverse text from the internet, books, and other sources. They employ deep learning techniques, particularly transformers \cite{vaswani2017}, which allow them to manage context over long passages of text effectively. This ability to maintain context makes them proficient in tasks such as translation, summarization, or question answering.

Moreover, LLMs have shown remarkable zero-shot, one-shot, and few-shot learning capabilities \cite{brown2020language}. This means they can perform tasks they were not explicitly trained on, given minimal examples or instructions. For instance, GPT-3 can write poetry, generate code, and even perform some reasoning tasks with little to no specific training.

\subsection{Techniques for LLMs}

LLMs leverage a variety of techniques to enhance their performance and usability. Some of the most notable techniques include:

\noindent\textbf{Prompt Engineering:} Crafting precise prompts guides the LLM in generating desired responses. Even slight variations in prompts can significantly influence the model's behavior and output, tailoring it to specific needs. Prompt engineering is essential for optimizing LLM performance across various applications \cite{reynolds2021prompt}.

\noindent\textbf{Chain of Thought:} This technique guides the model to generate logical steps leading to the final answer, breaking down complex problems for better reasoning and interpretability \cite{wei2022chain}. It has evolved from user-driven to model-driven CoT, demonstrating remarkable capabilities. Combined with technical ingenuity, this shift has enabled models like DeepSeek \cite{deepseek2024} to achieve outstanding performance.

\noindent\textbf{Few-Shot Prompting:} Generalizing from a small number of examples provided within the input prompt. This technique enables the model to infer patterns and apply them to new, unseen data \cite{brown2020language}.

\noindent\textbf{Fine-Tuning:} Training a pre-trained LLM on a specific dataset related to a particular task or domain. This adapts the LLM to specialized tasks by learning the nuances and specifics of the target domain \cite{howard2018fine}.

\noindent\textbf{Function Calling:} This technique leverages a structured JSON schema, allowing LLMs to interact with external codebases in an API-like manner. By specifying tools or functions in JSON format, users can instruct the model to execute tasks or access external data sources. For example, retrieving weather data might involve a function call like:

\begin{verbatim}
Task: Retrieve weather data for London using the function call.

Output:
{
    "function": "getWeatherData", "parameters": { "location": "London" }
}
\end{verbatim}

\subsection{Limitations of LLMs}

The biggest challenge facing LLMs is a phenomenon known as hallucinations. First introduced in the context of machine translation in 2018, the term refers to outputs that are entirely disconnected from the input, often incorrect or nonsensical \cite{lee2018hallucinations}. Despite their advanced capabilities, LLMs occasionally produce responses that lack grounding in the provided input or real-world facts. This issue poses significant challenges for applications requiring high accuracy and reliability. Ongoing research aims to better understand the causes of hallucinations and develop effective strategies to mitigate them \cite{ji2023hallucinations}.

At the same time, language has long been regarded as central to human reasoning, as argued by Wittgenstein and Piaget \cite{wittgenstein1953philosophical, piaget1926language}. Given this, the remarkable linguistic abilities of modern LLMs raise an important question: Could mastering language be the key to achieving artificial general intelligence? Motivated by this, this work explores pushing LLM capabilities beyond mere conversation.

\section{Kerbal Space Program (KSP)}

Kerbal Space Program (KSP)\footnote{\url{https://store.privatedivision.com/game/buy-kerbal-space-program-ksp}} is a space flight simulation video game developed by the Mexican studio Squad, released in 2015. Although it is a game, it can be used as a simulation environment with the addition of \textit{mods} that add new features such as more realistic physics.

The scope of KSP includes various spacecraft missions involving celestial bodies and maneuvering with goals such as escaping the atmosphere, reaching another planet, or intercepting a space object \cite{kerbal2014}. An interesting feature of this game is the ability to create missions that can be shared with other players. All these features convert KSP into a wonderful educational tool and simple work environment for initial testing of different space missions. 

Although KSP does not offer a perfect simulation of reality, it has been praised for its accurate orbital mechanics, even forming a partnership with NASA that elevated its status beyond just a game\footnote{\url{https://www.polygon.com/features/2014/1/27/5338438/kerbal-space-program}}. The simulation environment is constrained to a two-body problem and is limited to a small number of planets, most commonly just an earth-shaped planet called Kerbin\footnote{\url{https://kerbalspaceprogram.fandom.com/wiki/Kerbin}}.

\subsection{Kerbal Space Program Differential Games}
\label{sec:kspdg}

The KSPDG Challenge \cite{kspdg} facilitates a gymnasium and code for implementing agents into Kerbal Space Program (KSP). An agent is ``a software entity that performs tasks autonomously by perceiving its environment through sensors and acting upon that environment through actuators'' \cite{russell2016ai}.

In KSPDG, agents control a spacecraft's movement in all three rotational axes (yaw, pitch, and roll) using the thrust engines. Actions are expressed in the spacecraft's reference frame and include thrust magnitude for each axis and duration of the applied thrust. Although KSPDG allow numerical thrust magnitudes, the LLM agents presented in this study use discrete thrust values. KSPDG includes three scenarios:
\begin{itemize}[left=0pt]
    \item \noindent \textbf{Pursuer-evader (E1-E4)}: The agent controls the pursuer. The main objective is to minimize the distance between the pursuer and the evader. For different scenarios in the Pursuer-Evader game, the evader’s initial orbit remained constant across all scenarios while the pursuer’s initial orbit varied. The pursuer and the evader have identical vehicle parameters. Participants were evaluated on metrics such as distance between pursuer and evader (m), speed at closest approach (m/s), pursuer fuel usage (kg), and time elapsed (s). The main differences between scenarios E1, E2, E3, and E4 lie in the evasive maneuvers implemented by the Evader vessel. This study focuses on E3 which employs a structured evasive strategy, where the Evader vessel activates full thrust to escape when the Pursuer is within a specified distance threshold.
    \item \textbf{Target Guarding (Lady-Bandit-Guard):} Agents control a bandit spacecraft to stay close to a lady spacecraft while avoiding a guard spacecraft. Scenarios vary based on initial orbits and vehicle capabilities that will not be covered in this study.
    \item \textbf{Sun-Blocking:} Agents aim to position a spacecraft between an evader and the sun. Scenarios vary according to initial orbits and evader strategies that will not be covered in this study.
\end{itemize}

\begin{figure}[H]
  \centering
  \includegraphics[width=0.7\columnwidth]{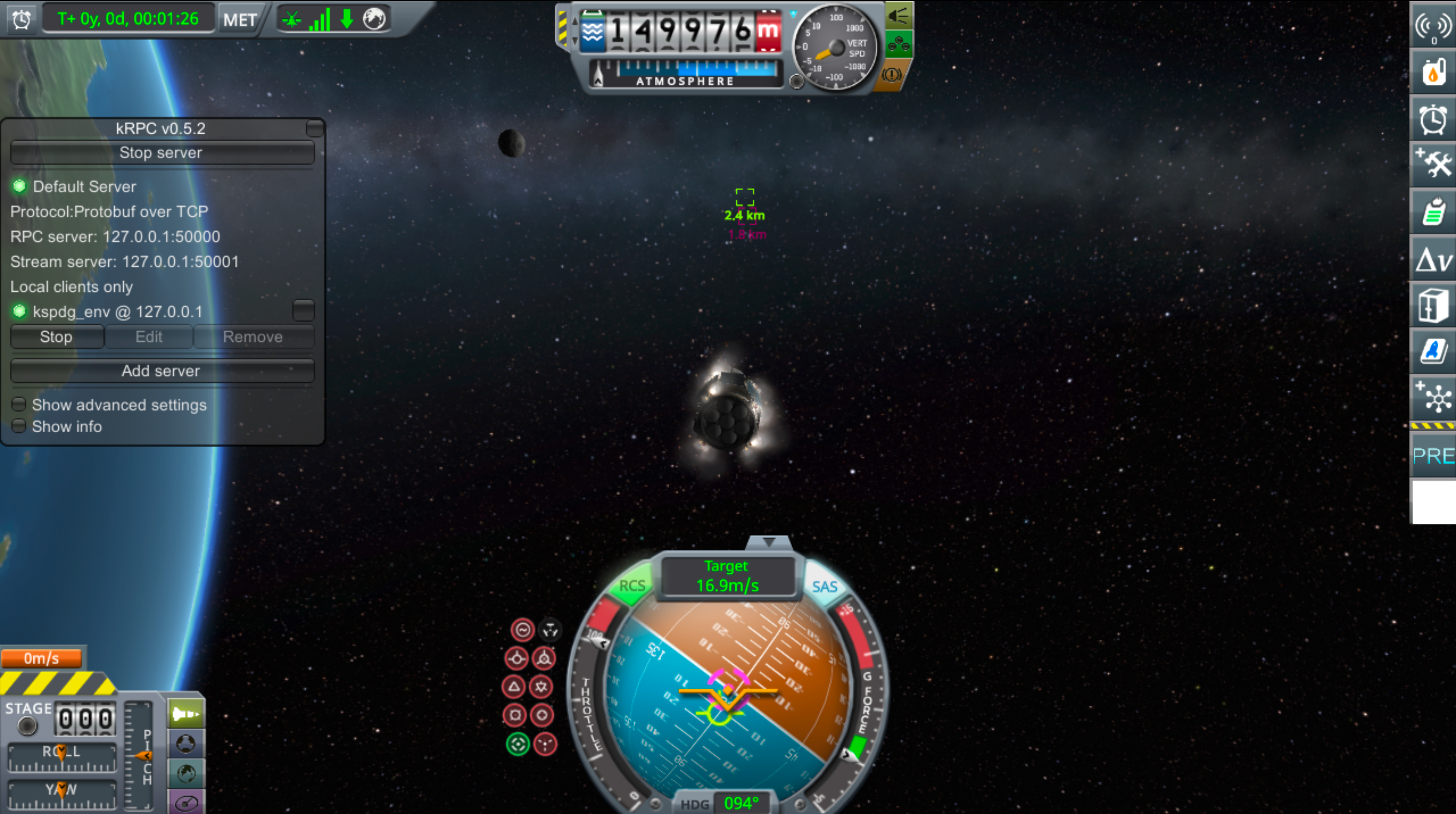}
  \caption{Agent’s in-game viewpoint during a KSP mission.}
  \label{fig:kspdg}
\end{figure}

\subsection{Navigation}
\label{sec:navigation}

As illustrated in \cref{fig:kspdg}, KSPDG provides a third-person perspective for navigation, incorporating multiple visual cues on a dashboard. The Navball, a spherical navigation display positioned at the bottom center of the screen, contains critical information necessary for in-spaceflight decision-making. A comprehensive explanation of in-space terminology and the markers used in KSP can be found in publicly available documentation (e.g., \cite{KSPNavball})

\section{GPT as a Spacecraft Operator}

The interaction with the Language Model (LLM) is primarily based on prompts. Prompts are the inputs provided to the model to elicit a desired response.

\subsection{GPT Data}
\label{sec:gpt-data}

For their GPT models, OpenAI uses a specific fine-tuning format\footnote{\url{https://platform.openai.com/docs/api-reference/chat}}. This format consists of a JSONL file with mandatory and optional parameters. Each entry includes a role: ``user'' for input prompts, ``system'' for system instructions, and ``assistant'' for model responses. This structure enables augmented chat generations for few-shot prompting \cite{touvron2023llama}.

KSPDG outputs \textit{observations} for each state of the mission. These \textit{observations} include the time elapsed (s), the vehicle mass (kg), the vehicle propellant (kg), positions for both pursuer and evader (x,y,z), as well as their velocities in the reference frame of the celestial body\footnote{The celestial body reference frame is a coordinate system centered on a celestial body (Kerbin) and rotates with it, while ``vessel up'' refers to the upwards direction in the vessel reference frame that is the coordinate system which is centered and oriented on the vessel under control} (m/s). Some of these observations may be of any relevance or not considered by the LLM.

At the time of the KSPDG challenge, the OpenAI API did not support maintaining context between calls. To address this limitation, a sliding window with zero padding was used in some experiments. This technique involves prepending the user prompt with the last n conversations with the LLM where n is the window size using zero padding if required \cite{beltagy2020longformer}.

Instead of using special tokens, some keywords were used to distinguish between the user and the assistant during chat interactions. Specifically, GPT identifies the user as \textit{HUMAN} and the model as \textit{ASISSTANT}.


\subsection{Prompt Engineering}
\label{sec:prompt_engineering}

Prompt engineering involves designing the input text for better comprehension by the LLM. We systematically generate multiple variations, run simulations in KSPDG, and select the best-performing prompts. Optimal prompts include concise explanations of the system, state observations, and mission goals.

Basic prompt engineering does not address the arithmetic limitations of LLMs when handling large numbers\footnote{This statement applies specifically to LLMs that have not been trained with explicit reasoning modules or reinforcement learning for arithmetic tasks\cite{deepseek2024, openai2024o1}}. Performing prompt augmentation, we enhance the observation space by computing the prograde vector to help the model achieve its mission.

The prograde vector is fundamental in orbital mechanics; it indicates the direction of movement of the spacecraft relative to a reference point. In the KSPDG environment, our spacecraft's nose always points toward the target, which serves as the reference for calculating the prograde. It is computed as follows:
\begin{equation}
\mathbf{p} = \mathbf{R}^{-1} \frac{\mathbf{v}_p - \mathbf{v}_e}{\|\mathbf{v}_p - \mathbf{v}_e\|}
\end{equation}

where:
\begin{itemize}
    \item $\mathbf{R}$: Rotation matrix that transforms coordinates from the vessel to the celestial body reference frame.
    \item $\mathbf{v}_p, \mathbf{v}_e$: pursuer's and evader's velocity in the celestial body reference frame.
\end{itemize}

The Rotation Matrix \(\mathbf{R}\) is defined as:
\begin{equation} 
\mathbf{R} = \begin{bmatrix} \mathbf{e}_r \mathbf{e}_f \mathbf{e}_u \end{bmatrix} \end{equation} where $\mathbf{e}_r, \mathbf{e}_f, \mathbf{e}_u$ are the unit vectors, in the reference frame of the celestial body, along the pitch (radial), roll (forward) and yaw (up) axes of the vessel.

\begin{equation}
\mathbf{e}_r = \mathbf{e}_f \times \mathbf{e}_u
\end{equation}

\begin{equation}
\mathbf{e}_f = \frac{\mathbf{p}_e - \mathbf{p}_p}{\|\mathbf{p}_e - \mathbf{p}_p\|}
\end{equation}

\begin{equation}
\mathbf{e}_u = \frac{\text{vessel\_up}}{\|\text{vessel\_up}\|}
\end{equation}

where:
\begin{itemize}
    \item $\mathbf{p}_p, \mathbf{p}_e$: pursuer's and evader's position in the celestial body reference frame.
    \item $vessel\_up$: refers to the predefined orientation of the spacecraft's top side relative to the celestial body.
\end{itemize}

\subsection{Few-shot Prompting}
\label{sec:few_shot}
The LLM often fails to perform the function call correctly in its first response, causing subsequent errors. To mitigate this, we use a hand-made example, similar to a one-shot prompt \cite{brown2020language}, including reasoning and a valid function call. This reduces erroneous function calls and improves subsequent responses.

In our previous work \cite{rodriguez-fernandez2024language}, we employed both a rule-based approach and Chain of Thought (CoT) prompting. While the rule-based approach improved performance, it significantly reduced generalization. In this study, we chose to retain only CoT, as it offers greater flexibility while maintaining strong performance. 

Using the CoT approach, the quality of the output is heavily reliant on the model's capacity for reasoning. Therefore, we used the most advanced version available for the GPT-3.5 model\footnote{The specific model version used in this CoT study is gpt-3.5-turbo-0125, which was the latest model available.}

\begin{figure*}[htbp]
\centering
\fbox{
\begin{minipage}{\textwidth}
\texttt{You operate as an autonomous agent controlling a pursuit spacecraft. Your goal is to apply
throttles to capture the evader given the positions and velocities of the pursuer and evader in celestial body reference frame and the direction of pursuer's velocity relative to evader or prograde. Throttles must be given
in your vessel reference frame wherein the x axis points to the right, the y axis points towards the target and
the z axis points upwards. The maximum throttle is 1. Reason step-by-step.
After reasoning call the perform\_action function.}
\end{minipage}
}
\caption{System prompt used for eliciting Chain of Thought}
\label{fig:system_prompt-CoT}
\end{figure*}

\begin{figure*}[htbp]
\centering
\fbox{
\begin{minipage}{\textwidth}

\textbf{USER}: Given these observations \{\textit{KSPDG environment observations from a past action}\}, 
what is the best throttle to capture evader?

\textbf{ASSISTANT}: The x coordinate of prograde is positive, indicating that pursuer is moving to the right. 
The y coordinate of prograde is negative, indicating that pursuer is approaching. 
The z coordinate of prograde is negative, indicating that pursuer is moving down. 
To capture the evader, we should move in the opposite direction in the x axis, 
towards the target in the y axis, and in the opposite direction in the z axis. 
This means we should apply throttles to move left, forward and up. Therefore we should call 
perform\_action(\{``ft'': ``forward'', ``rt'': ``left'', ``dt'': ``up''\}). \newline

Now answer the following question: \newline

Given these observations \{\textit{Current KSPDG environment observations}\}, 
what is the best throttle to capture evader?. \textbf{Reason step-by-step.}
\end{minipage}
}
\captionof{figure}{Sample user prompt with Chain-of-Thought (CoT) reasoning}
\label{fig:user_prompt_CoT}
\end{figure*}

Figure \ref{fig:system_prompt-CoT} introduces key terms (\textit{prograde}, \textit{reference frame}, \textit{celestial body}), essential for understanding the user prompt. Figure \ref{fig:user_prompt_CoT} illustrates an example of a user prompt using CoT which was crafted from the experience of previous interactions.

\subsection{Fine-tuning}
\label{sec:fine_tuning}
One notable characteristic of real-time control tasks is the need for rapid responses. OpenAI's ChatGPT completion API offers fine-tuning\footnote{GPT-3.5 Turbo fine-tuning and API updates: \url{https://openai.com/blog/gpt-3-5-turbo-fine-tuning-and-api-updates}} for several advantages, including improved result quality, increased training capacity, token savings, and reduced latency. We concluded that minimizing latency is critical in an agentic environment, as our agent may be unable to correct piloting errors due to the dynamics of orbital mechanics, making precise adjustments challenging.

We also addressed the scarcity of training data by developing a script to record human gameplay in KSP and convert it into sequential data for training a model to imitate human-like spacecraft maneuvering. The fine-tuning dataset consists of observations and thrust actions recorded every 0.5 seconds during gameplay. Thrust actions are logged as discrete labels relative to the spacecraft's reference frame (e.g., `backward,' `forward,' `left,' `right,' `down,' `top'). For instance, the term `forward' is mapped to a value of 1 in the forward thrust parameter, indicating maximum thrust.

To evaluate the performance of fine-tuning we run the following experiments sequentially, using \textit{GPT-3.5-turbo-1106}\footnote{As of February 2024, following the conclusion of the KSPDG challenge, alternative models available with fine-tuning capabilities include \textit{gpt-3.5-turbo-0613, babbage-002, davinci-002, and gpt-4-0613}. However, these demonstrated inferior performance in comparison.} as LLM baseline:
\begin{itemize}[left=0pt]
    \item \textit{Baseline LLM}: Standard ``gpt-3.5-turbo'' model with a generalized system prompt for Pursuer-Evader (PE) scenarios in the KSPDG challenge.
    \item \textit{Fine-tuning}: Trained in a single human gameplay log (314 user-assistant pairs) with default hyperparameters.
    \item \textit{+ Hyperparameter tuning}: Adjusting GPT-3.5 fine-tuning hyperparameters \cite{openai2023hyperparameters} improved performance. Lowering the learning rate multiplier to 0.2 enhanced exploration and convergence.
    \item \textit{+ System prompt}: Including a refined system prompt in training to balance clear instructions without over-constraining model behavior.
    \item \textit{+ Two train gameplays}: Adding another gameplay log (total 647 user-assistant pairs) to analyze learning improvements with more data.
\end{itemize}

GPT-4 and GPT-4o were unavailable during this study and are expected to perform better, as demonstrated in a related study \cite{carrasco2025visuallanguage}.

\section{LLaMA as a Spacecraft Operator}

Right after the end of the KSPDG Challenge, there was an interesting approach left to experiment with. While designing and developing the agents, the limited interaction with the OpenAI API and the high cost of it led us to find open-sourced LLMs that could be tweaked and worked with more data and more domain knowledge.

LLaMA is a family of \textit{large language models} released by Meta as a collection of foundational language models. The Hugging Face community, which also offers the \textit{transformers} library \cite{wolf2019huggingface}, provides a platform for hosting an AI community where a plethora of different LLM artifacts, such as open-source models and specific datasets, can be found. For our work, we utilized LLaMA-3-8B, the smallest model in the LLaMA 3 family, and their instruct version, which is post-trained for chat interactions.

In these LLaMA experiments we were determined to solve some of the problems that arose during GPT experiments. It is obvious that the final result of the model may badly generalize, even when including Chain of Thought (CoT) which clearly decreases this problem although not perfect. The goal of our LLaMA research is to build on the promising fine-tuning results observed in GPT, aiming for improved performance through the development of a more adaptable and robust model.

Given the additional data, the discrete labels relative to the spacecraft's reference frame included a new label for no action, called ``none'', which will be useful for situations when no action is better than an action. The none action is situational and was very difficult to learn with limited data files in GPT. The reference frame for this LLaMA fine-tuning will be the same as GPT's.

\subsection{LLaMA Data}

In an LLM, the tokenizer is typically unique to each model. This study leverages LLaMA 3, an open source model released in 2023 \cite{touvron2023llama}, to develop an autonomous pilot agent in Kerbal Space Program (KSP). The LLaMA 3 tokenizer includes special tokens (e.g., \texttt{<|begin\_of\_text|><|end\_of\_text|>}) that we used. For training, all files were converted to Alpaca \cite{stanford2023alpaca}.

\begin{lstlisting}[
    language=json,
    caption=JSON Snippet in Alpaca Format.,
    label=lst:alpaca_example,
    basicstyle=\ttfamily\small,
    frame=single,
    breaklines=true,
    columns=flexible,
    xleftmargin=0.025\textwidth,
    xrightmargin=0.025\textwidth
]
{
    "instruction": "\nHUMAN: Given these observations ...",
    "output": "The best throttle is ...",
    "system": "You operate as an autonomous agent ...",
    "history": []
}
\end{lstlisting}

Our LLaMA datasets utilized for training were entirely in Alpaca format, comprising $\sim100$ gameplays.

Additionally, we extracted flight logs from Navball bot, a prograde-alignment system designed to minimize thrust and corrections, and converted them into a synthetic dataset with dynamic chain-of-thought prompts for training all LLaMA models. For more information about this bot see \cref{sec:algorithms-appendix}.

\subsection{Few-shot Prompting}

Before fine-tuning, a few-shot prompting with CoT was designed to test the baseline LLaMA 3 model capabilities. The prompt can be seen in \ref{sec:prompts}.

\subsection{Optimization techniques}

The AI research may be well bottlenecked due to the requirements needed to train any state-of-the-art model. During our LLaMA research, we utilized a workstation equipped with a Threadripper CPU\footnote{\url{https://www.amd.com/en/products/processors/workstations/ryzen-threadripper.html}} and five RTX 4090 GPUs\footnote{\url{https://www.nvidia.com/es-es/geforce/graphics-cards/40-series/rtx-4090/}}.

To maximize efficiency and improve the training process, we employed a range of optimization techniques and tools:

\begin{itemize} [left=0pt]
    \item \textbf{LoRA \& DORA}: Reduce the number of trainable parameters by factorizing weight updates into low-rank and orthogonal low-rank matrices, enhancing efficiency and optimization \cite{hu2024lora, liu2024doraweightdecomposedlowrankadaptation}.
    \item \textbf{Quantization}: Initially applied to reduce model size and inference time, but was discarded due to performance degradation \cite{nagel2021white}.
    \item \textbf{LLaMA Factory}: LLaMA-Factory \cite{zheng2024llamafactory} is a project that serves as a LLaMA framework which provides a comprehensive set of tools and scripts for fine-tuning, exporting, serving and benchmarking LLaMA models.
    \item \textbf{Flash Attention (FA) and FA2}: Optimize memory and computation in attention mechanisms, reducing training and inference latency \cite{dao2022flashattentionfastmemoryefficientexact, dao2023flashattention2fasterattentionbetter}.
\end{itemize}
\subsection{Fine-tuning}
\label{sec:llama_fine-tuning}

To evaluate the performance of fine-tuning, we conducted the following sequential experiments using LLaMA-3 as the LLM baseline:

\begin{itemize}[left=0pt]
    \item Baseline LLaMA: This is the standard \textit{LLaMA-3} model used without any fine-tuning. It is programmed with a generalized system prompt designed for the Pursuer-Evader (PE) scenarios within the KSPDG challenge. This prompt includes the key aspects of rendezvous missions.
    \item Fine-tuning 10 documents: Fine-tuning LLaMA with 10 navball agent logs. For testing the LLaMA capabilities with a small dataset.
    \item Fine-tuning 10 documents with sliding window of 3: Fine-tuning LLaMA with 10 navball agent logs using a sliding window of 3 actions. This model attempts to predict the next three actions based on the previous ones.
    \item Fine-tuning with 10 documents and look ahead: Fine-tuning LLaMA with 10 Navball agent logs using a look-ahead approach, where the agent outputs a series of sequential actions. This model is trained to predict the next three actions.
    \item Fine-tuning 25 documents: After increasing the training data to 25 navball agent logs. To test how LLaMA may improve given a medium dataset.
    \item Fine-tuning 50 documents: After further increasing the training data to 50 navball agent runs. To test the LLaMA fine-tuning with a rich set of gameplays.
\end{itemize}

Having outlined the iterations of each experiment, we now detail the hyperparameters used for fine-tuning:

\begin{table}[H]
\centering
\begin{tabular}{ll}
  \hline
  \textbf{Hyperparameter} & \textbf{Value} \\
  \hline
  AdamW Optimizer & Yes \\
  Quantization & None \\
  Learning Rate & 1e-4 \\
  Learning Rate Scheduler & Cosine \\
  LoRA Configuration & $\{r=16,\alpha=8,d=0.05\}$ \\
  Flash Attention 2 & Enabled \\
  Epochs & 3 \\
  Dora & Enabled \\
  Trainable Layers & Last 2 layers \\
  Batch Size & 2 \\
  Gradient Accumulation Steps & 2 \\
  \hline
\end{tabular}%
\caption{Hyperparameters used for fine-tuning LLaMA}
\label{tab:llama-hyperparameters}
\end{table}

Gradient accumulation steps help compensate for the small batch size of 2, chosen due to GPU limitations and to balance convergence speed with generalization. LoRA parameters are optimized for stability, with only the top two layers trainable. Flash Attention 2 improves efficiency, while AdamW and a cosine learning rate scheduler facilitate smooth convergence. Dora dynamically adjusts the training parameters to further improve optimization.

\section{Results}

In this section, we evaluate the performance of various models. We will also use the metrics designed for the KSPDG Challenge \cite{kspdg} as well as some of our own metrics and evaluations.

\subsection{Training Metrics}

The accuracy of the inferred actions is assessed using a discrete evaluation metric that determines whether the predicted action matches the ground truth. Specifically, the overall action accuracy is defined as

\begin{equation}
\text{Acc}_{\text{action}}(\mathbf{a},\hat{\mathbf{a}}) =
\begin{cases}
1, & \text{if } \mathbf{a} = \hat{\mathbf{a}},\\[6pt]
0, & \text{otherwise,}
\end{cases}
\end{equation}

where \(\mathbf{a}\) is the ground truth action vector and \(\hat{\mathbf{a}}\) is the inferred action.
Consequently, we compute a custom cross-entropy loss using the logistic loss function, assuming that the LLM agent produces a one-hot encoded prediction corresponding to \(\hat{\mathbf{a}}\).

Next, the performance is quantified using KSPDG's scoring function defined by
\begin{equation}
\begin{aligned}
\text{Score} &= \left( 0.1 \cdot d \right)^{2.0} + \left( 0.5 \cdot v \right)^{1.5} \\
&\quad + \left( 0.1 \cdot f \right)^{1.25} + \left( 0.01 \cdot t \right)^{1.0}
\end{aligned}
\end{equation}

where \(d\), \(v\), \(f\), and \(t\) denote distance (m), velocity (m/s) fuel usage (l) and time (s), respectively. In each term, the component is scaled by a factor (e.g., \(0.1\), \(0.5\), etc.) and raised to an exponent (e.g., \(2.0\), \(1.5\), etc.). More generally, the scoring function is given by

\begin{equation}
\text{Score} = \sum_{c} (s_c \cdot c)^{w_c},
\end{equation}

with \(s_c\) and \(w_c\) representing the scale and exponent for component \(c\), respectively.

\subsection{Access to Experiments and Models}

All experiments carried out during this investigation have been tracked using Weights \& Biases (wandb). The projects are publicly accessible and provide detailed logs, metrics, and visualizations of the training processes. Additionally, the models developed and used in this research are hosted on Hugging Face. Each repository contains specific models and datasets relevant to different aspects of the research.

Details of the specific projects, including experimental runs, fine-tuning processes, and final versions of the models, including various versions of LLaMA models, can be found in the appendix. (See \ref{sec:detailed-wandb} and \ref{sec:detailed-huggingface})

\subsection{GPT Training Results}

Our GPT fine-tuning approach involved small datasets and a few runs due to the limitations that the API offers given its proprietary nature. Nonetheless, a minor fine-tuning was conducted using a small dataset to validate how well GPT adapts to the data and to examine the hypothesis that increased data ingestion leads to improved performance.

One key hyperparameter to highlight is the Learning Rate Multiplier (LRM). While OpenAI’s default settings are generally effective, the default LRM value of two proved problematic, causing rapid convergence and high overfitting. As summarized in \cref{tab:model-losses-gpt}, our incremental experiments with GPT showed improvements in training loss but failed to yield similarly favorable validation loss outcomes. We attribute this discrepancy to the limitations of fine-tuning a non-locally hosted, closed-source model, which restricts flexibility. This constraint ultimately motivated our shift toward open-source LLMs, allowing for greater control over the training process.

\begin{table}[H]
\centering
\begin{tabular}{l|c|c}
    \hline
    \textbf{Model Name} & \textbf{Training Loss} & \textbf{Validation Loss} \\
    \hline\hline
    \multicolumn{3}{c}{\textbf{Final Models}} \\
    \hline\hline
    Default LRM          & \num{0.47663} & - \\
    + LRM 0.2            & \num{0.02292} & \num{0.02772} \\
    + system prompt      & \num{0.18502} & \textbf{\num{0.00519}} \\
    + 2 human gameplays  & \textbf{\num{0.00013}} & \num{0.43595} \\
    \hline
\end{tabular}
\caption{Training and Validation Loss for Each GPT Model}
\label{tab:model-losses-gpt}
\end{table}

\begin{figure}[ht]
  \centering
  \includegraphics[width=0.9\columnwidth]{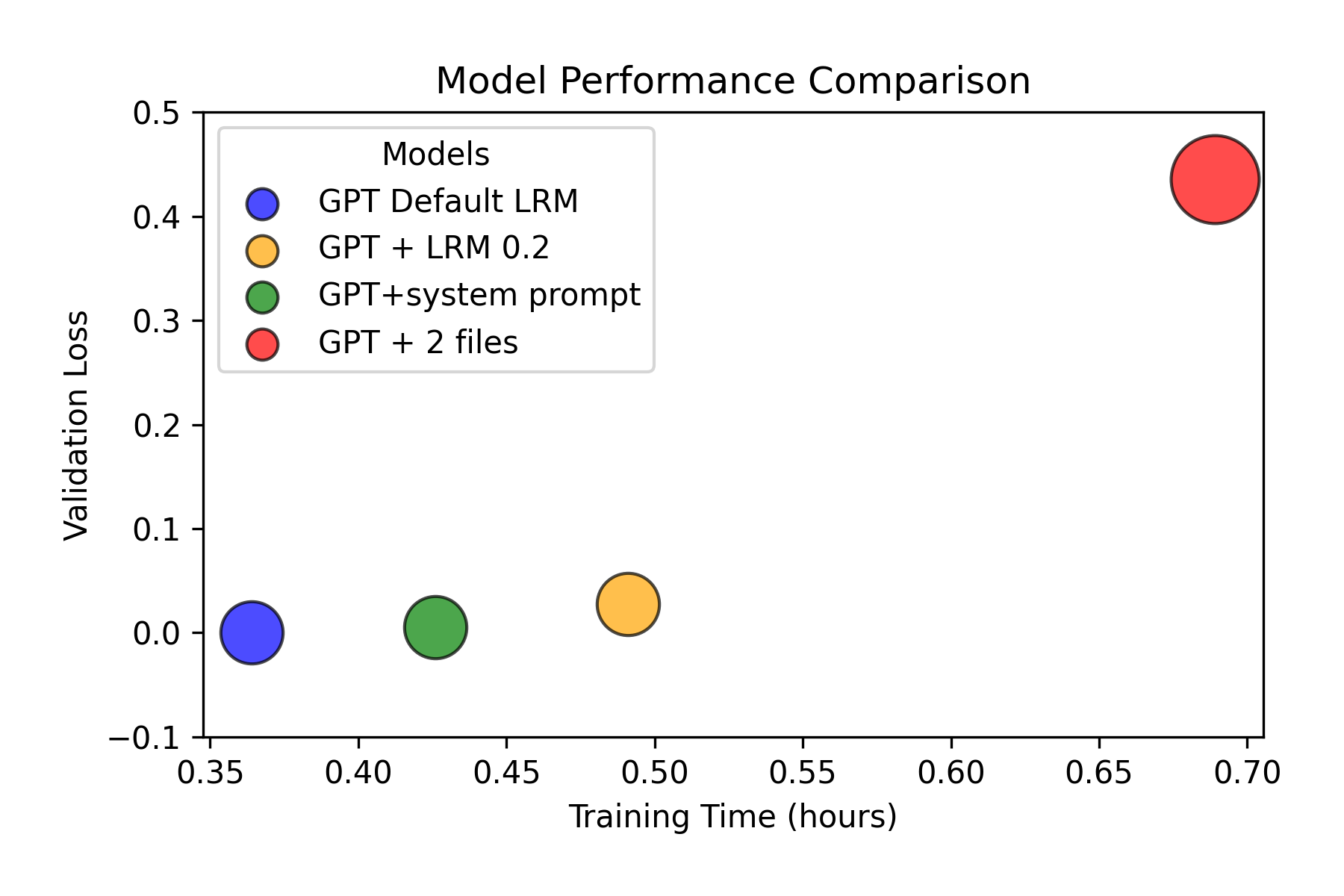}
  \caption[Figure of Sliding Window.] {Model Performance with Final Hyperparameters (GPT). Circle size represents training dataset size. GPT Default LRM lacked a validation set, marked as zero.}
  \label{fig:loss_gpt_training}
\end{figure}

\subsection{LLaMA Training Results}

Our fine-tuned LLaMA models can be categorized into two distinct groups: the experimental models, which were trained during a phase where we experimented with variations of hyperparameters, and the final models, which incorporate all the final hyperparameters presented in \cref{tab:llama-hyperparameters}.

\begin{figure}[H]
  \centering
  \includegraphics[width=0.9\columnwidth]{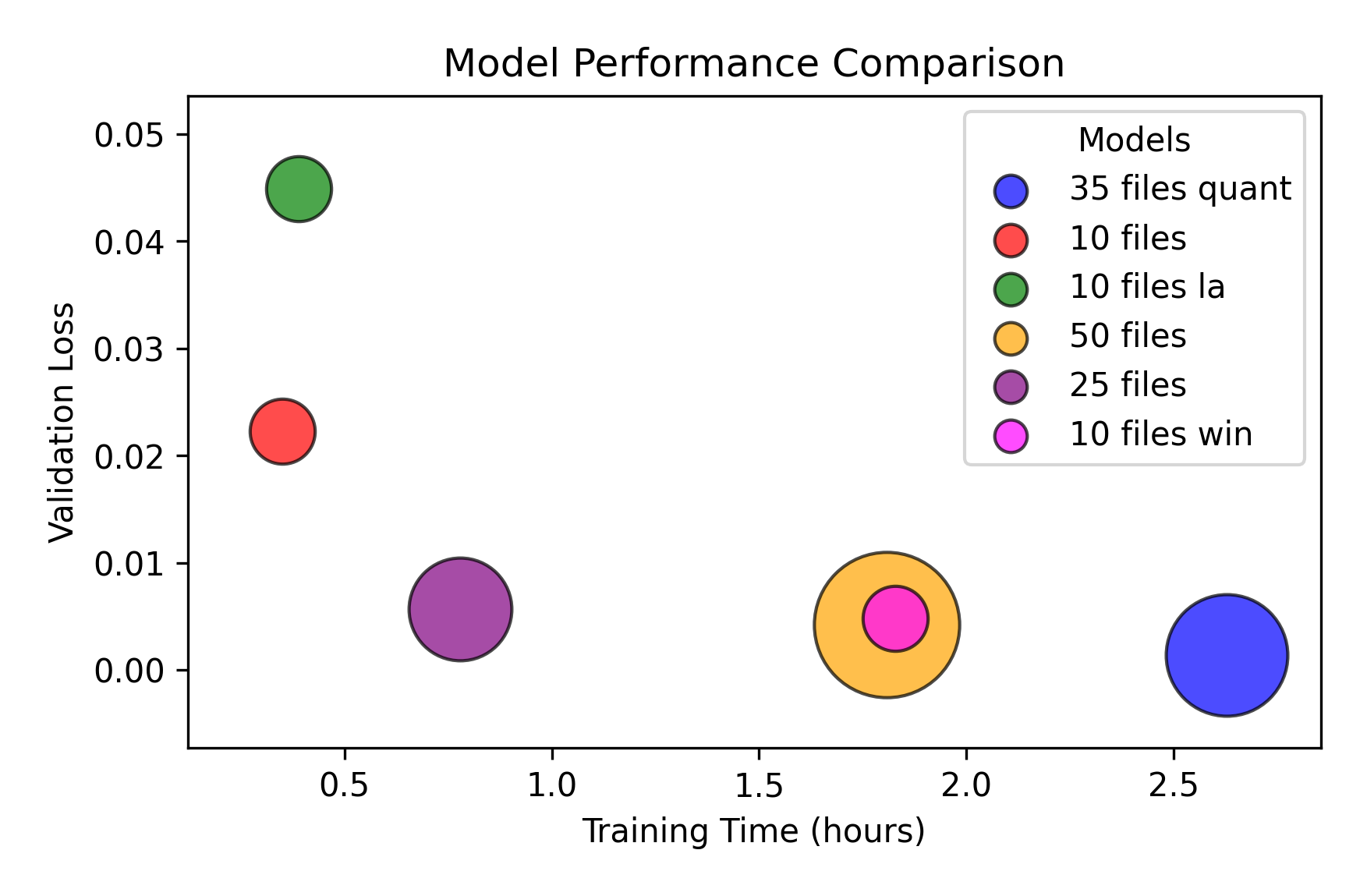}
  \caption[Figure of Sliding Window.] {Comparison of Model Performance with Final Hyperparameters and Various Dataset Sizes (LLaMA). ``Quant'' indicates models with quantized parameters, ``la'' refers to look ahead, and ``win'' denotes models trained with a sliding window dataset approach. The size of the circle refers to the size of the dataset.}
  \label{fig:model-performance-llama}
\end{figure}

When comparing the experimental models to the final versions, we observed a significant improvement on the order of one to two magnitudes. Table \ref{tab:model-losses-llama} shows the losses for all models. As also shown in \cref{fig:model-performance-llama}, models with larger datasets generally achieve better results. However, an exception is the model that uses a sliding window of 3, which, despite having fewer files, performs similarly to the 50-file model. This indicates that the sliding window approach is highly effective in enabling the model to learn from fewer gameplays. There was an experiment attempting to predict the next `x' steps of the spacecraft, referred to as the `look ahead' (la) method. However, it was not very effective and is therefore symbolically explained in the methodology.

\vspace{20 mm}

\begin{table}[H]
\centering
\begin{tabular}{l|c|c}
    \hline
    \hline
    \textbf{Model Name} & \textbf{Training Loss} & \textbf{Validation Loss} \\
    \hline
    \hline
    \multicolumn{3}{c}{\textbf{Experimental Models}} \\
    \hline
    Huggingface Models & \num{0.6829} & \num{0.6963} \\
    LLaMA-Factory Models & \textbf{\num{0.0284}} & \textbf{\num{0.0374}} \\
    \hline
    \hline
    \multicolumn{3}{c}{\textbf{Final Models}} \\
    \hline
    \hline
    LLaMA 35 files quant & \num{0.0151} & \textbf{\num{0.0014}} \\
    LLaMA 10 files la & \num{0.0573} & \num{0.0449} \\
    LLaMA 10 files win & \textbf{\num{0.0106}} & \num{0.0048} \\
    LLaMA 10 files & \num{0.0284} & \num{0.0223} \\
    LLaMA 25 files & \num{0.0161} & \num{0.0057} \\
    LLaMA 50 files & \textbf{\num{0.0104}} & \num{0.0042} \\
    \hline
\end{tabular}%
\caption{Training and Validation Loss for Each LLaMA Model}
\label{tab:model-losses-llama}
\end{table}

Throughout the chronology of all the different training runs we can see how the model is enhanced due to the gradual improvements, which demonstrates the effectiveness of the path chosen for these trainings.

All evaluations were conducted using LLaMA's 8B Instruct model, which was selected based on GPU memory capacity, inference speed considerations, and its foundational knowledge and format compatibility.

\subsection{GPT Performance}

Interestingly, while the OpenAI API for fine-tuning requires customization, it provides limited tools, especially compared to LLaMA. Hence, the effectiveness of GPT training is heavily dependent on the quantity and quality of data, together with certain adjustments (notably hyperparameters), with the LRM having the most significant impact.

The application of the Chain of Thought approach demonstrates a significant improvement in the generalization of spacecraft piloting techniques for the Pursuer-Evader problem as well as guiding the model to achieve a 0\% failure rate in execution. Results in Table \ref{tab:table_cot_1} indicate a consistent achievement of target proximity within 25 meters across varied scenarios, highlighting the approach's effectiveness in enhancing model precision and adaptability in complex rendezvous tasks. It is evident that this advancement notably enhances the model reasoning for its spacecraft operating capabilities.

\begin{table}[H] 
\centering
\resizebox{\columnwidth}{!}{%
    \begin{tabular}{l|cc|cc|cc|cc|c}
    \hline
    \multirow{2}{*}{\textbf{Method}} & \multicolumn{2}{c|}{\textbf{E1 (m)}} & \multicolumn{2}{c|}{\textbf{E2 (m)}} & \multicolumn{2}{c|}{\textbf{E3 (m)}} & \multicolumn{2}{c|}{\textbf{E4 (m)}} & \multirow{2}{*}{\textbf{Failure Rate}} \\
    \cline{2-9}
     & \textbf{Best} & \textbf{Avg} & \textbf{Best} & \textbf{Avg} & \textbf{Best} & \textbf{Avg} & \textbf{Best} & \textbf{Avg} & \\
    \hline
    Naive & 229.75 & 267.51 & 182.28 & 236.10 & 225.0 & 225.0 & 215.20 & 216.49 & - \\
    Baseline GPT & 36.42 & 269.22 & 245.76 & 295.09 & 292.59 & 328.20 & 305.02 & 329.78 & 37.5\% \\
    w/ CoT & \textbf{11.02} & \textbf{14.59} & \textbf{15.73} & \textbf{21.28} & \textbf{5.63} & \textbf{21.71} & \textbf{10.42} & \textbf{17.27} & \textbf{0.00\%} \\
    \hline
    \end{tabular}
}
\caption[Performance of the GPT agent w/ CoT]{Performance of the GPT agent (closest distance) with Chain of Thought (CoT) prompting across different Pursuer-Evader environments.}
\label{tab:table_cot_1}
\end{table}

Table~\ref{tab:fine_tune_performance} summarizes the performance of our fine-tuned agents. An error in calling a function causes the environment to terminate or the agent to freeze; these are considered failures. During all successful runs, we calculate the best and average closest approach distance. We compare these results against the agents used in SpaceGym \cite{10115968}. Specifically, we compare them against a naive agent, a Lambert model predictive control (Lambert-MPC) agent, an iLQGames agent \cite{fridovichkeil2020efficient}, and a Proximal Policy Optimization (PPO) agent, as benchmarked in the SpaceGym paper \cite{10115968}.We refer the reader to SpaceGym to read more about these agents.

As shown in the table, each fine-tuning experiment improves our agent’s performance while also reducing run variance. Notably, fine-tuning significantly decreases GPT-3.5 response latency, as demonstrated in Table~\ref{tab:table_cot_1}. This improvement is primarily due to the reduced number of output tokens generated by the fine-tuned models, as well as the high failure rate observed in the GPT baseline (Table~\ref{tab:table_cot_1}), where most failures stem from function call invocation errors.

Additionally, a key insight from both Table~\ref{tab:fine_tune_performance} and Table~\ref{tab:fine_tune_latency} is that fine-tuning yields progressively better results with continued refinements. This suggests that our fine-tuning strategy has yet to reach a performance plateau, indicating that incorporating more data could further enhance our agents' capabilities and unlock even greater improvements.

\begin{table}[h]
\centering
\begin{tabular}{l|ccc}
    \hline
    \textbf{Method} & \textbf{Best Dist. (m)} & \textbf{Avg Dist. (m)} & \textbf{Failure Rate} \\
    \hline
    Naive & 225.0 & 225.0 & - \\
    PPO & >1643 & 2346 & - \\
    iLQGames & >53.31 & 60.99 & - \\
    Lambert-MPC & >18.24 & 47.94 & - \\
    \hline
    Baseline GPT & 178.11 & 200.10 & 36.8\% \\
    Simple fine-tuning & 263.55 & 265.89 & \textbf{0.0\%} \\
    + Hyperparam tuning & 188.90 & 202.08 & 0.1\% \\
    + System prompt & 197.41 & 214.87 & \textbf{0.0\%} \\
    + Two train gameplays & 132.09 & 159.78 & 0.2\% \\
    \hline
    Human Gameplay & \textbf{5.97} & \textbf{6.25} & - \\
    \hline
\end{tabular}
\caption{Performance of GPT for each fine-tuning technique in meters.}
\label{tab:fine_tune_performance}
\end{table}

As shown in Table~\ref{tab:table_cot_1}, the GPT-3.5 baseline model's failure rate in KSP scenarios exceeds tolerable limits, while fine-tuned models reduce failures to nearly 0\%. Crucially, only with all incremental refinements do we surpass the baseline prompt-engineered model without chain of thought, underscoring the need for more training samples.

\begin{table}[H]
\centering
\resizebox{\columnwidth}{!}{
    \begin{tabular}{l|ccc}
    \hline
    \textbf{Method} & \textbf{Best Latency} & \textbf{Average Latency} & \textbf{Standard Deviation} \\ \hline
    baseline GPT & 759.69 & 840.42 & 83.85 \\
    simple fine-tuning & 977.32 & 987.43 & \textbf{15.08} \\
    + hyperparameter tuning & 749.02 & 831.30 & 49.02 \\
    + system prompt & 684.13 & 753.52 & 51.75 \\
    + two train gameplays & \textbf{468.98} & \textbf{557.49} & 89.54 \\ \hline
    \end{tabular}
}
\caption{Latency of GPT responses for each experiment in milliseconds.}
\label{tab:fine_tune_latency}
\end{table}

As illustrated in \cref{fig:trajectories_grid_plot}, there is a notable issue when comparing the current model with the GPT agent, specifically regarding overshooting. This problem arises from an incorrect prompt hint (see \cref{sec:fine_tuning}) to \textit{accelerate in the direction of the relative position of the evader}. Consequently, this forces the model to adopt an accelerate-only approach. This issue will be addressed in the subsequent LLaMA experiments.

\subsection{LLaMA Performance}

In contrast to the GPT experiments, the LLaMA experiments were specifically designed to achieve superior fine-tuning results. This focus is due to the inherent nature of an open-source model, that is intended to be enhanced and customized by its users.

The iterative process for all experiments is explained in \cref{sec:llama_fine-tuning}. Table~\ref{tab:llama-hyperparameters} summarizes the performance metrics of the LLaMA fine-tuning experiments. 

\begin{table}[H]
\centering
\resizebox{\columnwidth}{!}{
    \begin{tabular}{l|ccc}
    \hline
    \textbf{Method} & \textbf{Best Distance (m)} & \textbf{Average Distance (m)} & \textbf{Failure Rate} \\ \hline
    Naive & 225.0 & 225.0 & - \\
    PPO & >1643 & 2346 & - \\
    iLQGames & >53.31 & 60.99 & - \\
    Lambert-MPC & >18.24 & 47.94 & - \\
    \hline
    baseline LLaMA & 52.69 & 140.68 & 9.09\% \\
    fine-tune 10 files & 30.52 & 51.53 & \textbf{0.00\%} \\
    fine-tune 25 files & 13.54 & \textbf{29.44} & \textbf{0.00\%} \\
    fine-tune 50 files & \textbf{11.86} & 29.76 & \textbf{0.00\%} \\
    fine-tune 10 files win=3 & 23.08 & 40.03 & \textbf{0.00\%} \\
    \hline
    Navball (bot) & 34.34 & 36.43 & - \\
    \hline
    \end{tabular}
}
\caption[Performance of LLaMA for each fine-tuning technique in meters]{Performance of LLaMA for each fine-tuning technique in meters. The scenario used for these results is E3.}
\label{tab:llama_fine_tune_results}
\end{table}

The LLaMA results have exceeded our expectations. Not only did the model follow a robust prograde trajectory (see \cref{sec:trajectories}) but it also outperformed almost every other method in the KSPDG Challenge. While these results are second only to the Chain of Thought (CoT) approach, it could be argued that CoT ``hacks'' the language model via prompting, potentially undermining its intrinsic reasoning capabilities. It is noteworthy that all dataset files were collected using the navball bot, and the LLaMA fine-tuned model moderately outperforms the navball bot's performance.

It is important to discuss that the base LLaMA model achieves better results than the GPT model. However, this outcome is expected when considering that LLaMA-3 competes with GPT-4, rather than GPT-3 which is the model that was used in our GPT approach.

One technique utilized in these experiments but not in the GPT experiments due to limitations was the sliding window method (see \cref{sec:gpt-data}). As evidenced by the training and experimental results, despite having a significantly smaller data set, the sliding-window model competes with the top three best performing fine-tuned models.

Furthermore, we can see the latency results in \cref{tab:llama_latency}. The observed latencies indicate a significant delay that adversely affects the model's performance when utilized by an agent, especially when compared to our GPT models. Optimizing the model or upgrading the technical resources may lead to better results.

\begin{table}[h]
\centering
\resizebox{\columnwidth}{!}{
\begin{tabular}{l|ccc}
    \hline
    \textbf{Method} & \textbf{Best Latency} & \textbf{Average Latency} & \textbf{Standard Deviation} \\ \hline
    baseline LLaMA & 6449.16 & 8580.43 & 1175.69 \\
    LLaMA ft 10 files & 3305.15 & 3444.88 & 21.28 \\
    LLaMA ft 25 files & 3282.21 & 3316.89 & \textbf{21.19} \\
    LLaMA ft 50 files & 3418.97 & 3455.29 & 31.22 \\
    LLaMA ft 10 files win=3 & \textbf{3252.89} & \textbf{3292.44} & 38.93 \\
    \hline
\end{tabular}
}
\caption{Latency of LLaMA responses for each experiment in milliseconds.}
\label{tab:llama_latency}
\end{table}

Ultimately, as illustrated in the appendix in \cref{fig:trajectories_fine-tuned_llama}, the best-performing models exhibit a ``clunky'' movement pattern due to action delays. However, the overall distributions are quite similar. The navball bot should serve as the benchmark to evaluate the models' performance. 

\subsection{Scoring and Accuracy results}

Table \ref{table:ft_models_accuracy_and_cross-entropy} presents the loss and accuracy of the cross-entropy of various fine-tuned Llama and GPT models in scenario E3. Notably, the LLaMA models demonstrate excellent results, significantly outperforming the GPT models.

\begin{table}[H]
\centering
\begin{tabular}{l|c|c}
    \hline
    \textbf{Model} & \textbf{Accuracy} & \textbf{Cross-Entropy Loss} \\
    \hline
    Llama 10 files & 0.83 & 5.95 \\
    Llama 25 files & 0.80 & 7.27 \\
    Llama 50 files & \textbf{0.96} & \textbf{1.32} \\
    Llama 10 files win=3 & 0.95 & 1.65 \\
    GPT Default LRM & 0.049 & 34.27 \\
    GPT + LRM 0.2 & 0.20 & 28.76 \\
    GPT + system prompt & 0.37 & 22.55 \\
    GPT + 2 human gameplays & 0.26 & 26.69 \\
    \hline
\end{tabular}
\caption{Cross-entropy loss and Accuracy of Llama and GPT Fine-tuned Models in Scenario E3. These metrics were only used on trained models.}
\label{table:ft_models_accuracy_and_cross-entropy}
\end{table}

The results shown in \cref{table:ft_models_accuracy_and_cross-entropy} indicate that the Llama fine-tuning yields exceptional performance, with significantly higher accuracy and cross-entropy loss compared to the GPT models. It's important to note that the GPT models only had access to 2 files, which likely limited their performance. 

This performance extends beyond the training distribution, highlighting a remarkable outcome: the LLaMA models outperformed the navball bot that originally generated the training data. This unexpected result suggests that the LLaMA models were able to leverage their prior knowledge and reasoning capabilities to make more effective decisions, surpassing the baseline set by the navball bot. The discrepancy in accuracy underscores the potential of fine-tuned models to generalize beyond their training data, demonstrating a level of adaptability that was not initially anticipated.

The fine-tuned LlaMA Model with 50 files, achieved the highest accuracy and the lowest cross-entropy,  closely followed by the 10 files using sliding window model. This performance underscores the model's ability to generalize better from a larger dataset, significantly optimizing its decision-making process.

The results of the KSPDG scoring evaluation are presented in \cref{table:kspdg-scoring} below. Please note that best speed means best \textbf{approach} speed, this means the speed of the pursuer at the closest distance.

\begin{table}[htbp]
\centering
\small
\setlength{\tabcolsep}{2pt}
\resizebox{\columnwidth}{!}{
\begin{tabular}{l|c|c|c|c}
    \hline
    \textbf{Experiment} & \textbf{Best Score} & \textbf{Best Distance} & \textbf{Best Speed} & \textbf{Best Fuel Usage} \\
    \hline
    \hline
    \multicolumn{5}{c}{\textbf{GPT Experiments}} \\
    \hline
    \hline
    GPT Baseline                                    & 759.69 & 178.11 & 77.77 & 651.54 \\
    GPT Default LRM                         & 977.32 & 263.55 & 74.62 & \underline{220.84} \\
    GPT + LRM 0.2                         & 749.02 & 188.90 & 72.66 & 611.55 \\
    GPT + system prompt                   & 684.13 & 197.41 & 75.91 & \dashuline{235.49} \\
    GPT + 2 files                         & 468.98 & 132.09 & 73.58 & 267.73 \\
    \hline
    \hline
    \multicolumn{5}{c}{\textbf{LLaMA Experiments}} \\
    \hline
    \hline
    LLaMA Base Model                                & 387.55 & 140.68 & 50.76 & 390.77 \\
    LLaMA 10 files (optimized)            & 232.96 & 30.52  & 19.26 & 615.11 \\
    LLaMA 25 files (optimized)            & \dashuline{184.31} & \dashuline{13.54}  & \dashuline{18.01} & 436.02 \\
    LLaMA 50 files (optimized)            & 194.71 & \underline{11.86}  & 27.29 & 471.58 \\
    LLaMA 10 files win=3 (optimized)      & 212.17 & 23.08  & 24.83 & 462.08 \\
    \hline
    \hline
    \multicolumn{5}{c}{\textbf{Training Data Distribution}} \\
    \hline
    \hline
    Navball (bot)                       & \textbf{63.64}  & 34.34           & \underline{16.62}         & \textbf{123.19} \\
    Human gameplay                                  & \underline{72.39}      &  \textbf{5.97}  & \textbf{3.77}  & 1125.70\\
    \hline
\end{tabular}
}
\caption{Best Parameters and Scoring for Different Models. \textbf{Bold} indicates best, \underline{underline} indicates second best, and \dashuline{dashed underline} indicates third best.}
\label{table:kspdg-scoring}
\end{table}

Since the LLaMA fine-tuning took place after the conclusion of the KSPDG Challenge, scoring was not the primary focus of our approach. Instead, we prioritized optimizing the model to minimize the best distance, which is reflected in the results.

Additionally, the model's latency increases the likelihood of failing minor adjustments and prograde alignments, resulting in higher propellant consumption for corrections. Even so, the results for the other factors are far from bad.

For the GPT runs, although human gameplay is superior, we observe gradual improvements and better fuel efficiency across iterations. However, this efficiency arises because the model often ``decides'' not to move.

\subsection{Comparison of trajectories}
\label{sec:trajectories}

There were some worth noticing patterns that helped us improve the models and validate their reasoning during the KSP gameplays.

\begin{figure}[htbp]
    \centering
    \begin{subfigure}[b]{0.45\textwidth}
        \centering
        \includegraphics[width=\textwidth]{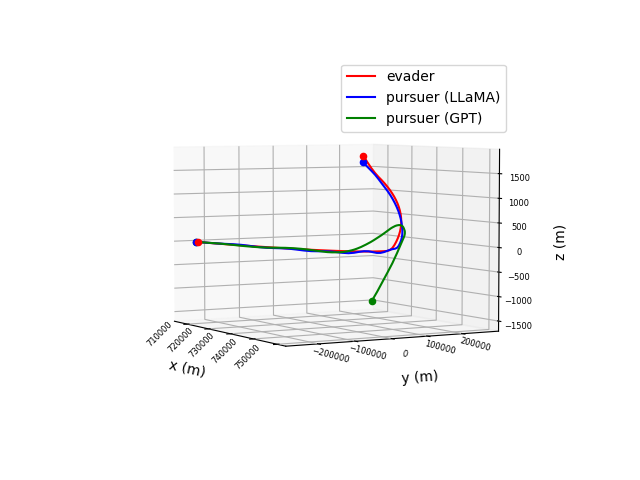}
        \caption{Fine-tuning trajectories}
        \label{fig:fine-tuning-trajectories-results}
    \end{subfigure}
    \hfill
    \begin{subfigure}[b]{0.45\textwidth}
        \centering
        \includegraphics[width=\textwidth]{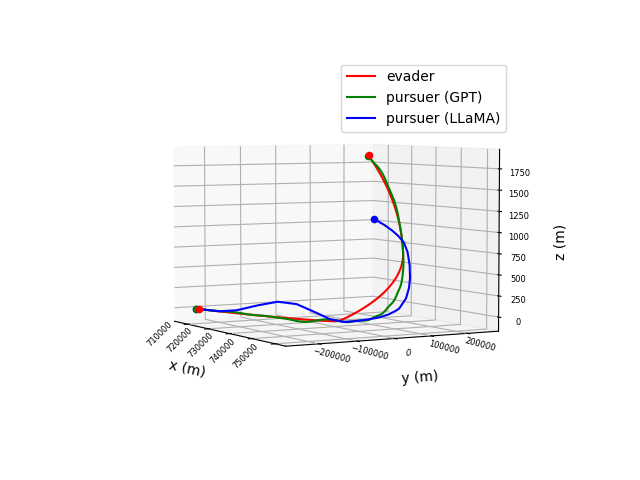}
        \caption{Few-shot prompting trajectories}
        \label{fig:few-shot-trajectories}
    \end{subfigure}
    \caption{Comparison of fine-tuning and few-shot prompting trajectories. The red line represents the trajectory of the evader.}
    \label{fig:3d-trajectories-results}
\end{figure}

The figures in \cref{fig:few-shot-trajectories} and \cref{fig:fine-tuning-trajectories-results} illustrate key points discussed earlier. The red line represents the trajectory of the evader.

The fine-tuning trajectories in \cref{fig:fine-tuning-trajectories-results} indicate that the data ingested by the model aids in understanding the problem and determining appropriate actions. However, these trajectories also show that the model's prior knowledge and reasoning still influence its performance. For instance, an incorrect hint, as depicted by the GPT trajectory (green line), deteriorates the model's performance and makes the agent `overshoot' (receding from the evader). In contrast, an ``agnostic'' prompt that does not dictate the model's reasoning can even enhance data-driven performance.

On the other hand, the few-shot prompting trajectories using the Chain of Thought technique are similar between GPT and LLaMA models. GPT performed slightly better due to its significantly lower latency.

\section{Conclusions \& Future Work}
\label{sec:conclusions&futurework}

We find the results very satisfying. The spacegym integration, alongside the orbit generation and agent integration, demonstrated that Kerbal Space Program can be a great, yet simple, alternative as a simulation engine.

The findings obtained within a brief training period of two to four months for model development and agent implementation provide strong evidence of the effectiveness of this approach as a robust and viable alternative. We successfully created an agent that could mimic and even improve upon ``expert'' data such as synthetically generated data or human gameplays.

Following this, there is no doubt that training a LLM can leverage prior knowledge and improve it for specific scenarios. The LLaMA agent, as an open-sourced model, was more complex to train but offered a plethora of tools, frameworks, and hyperparameters. These LLaMA models significantly outperformed the GPT ones, which achieved second place in the KSPDG Challenge, only behind a differential equations approach, and not too far from this said approach.

It is important to note that the base LLaMA model achieves better results than the GPT model. However, this outcome is expected when considering that the LLaMA-3 model competes with the GPT-4 model, rather than the GPT-3 model used in the original GPT approach.

Enhancing the model's optimization or upgrading the technical resources may lead to improved results, as these outcomes are highly hardware-dependent. In contrast, the GPT models were fully trained using OpenAI's powerful resources, which completely outweigh our local ones. Although the LLaMA model's results are already better, it is important to note that there is still room for further improvement.

Moreover, it is crucial to reflect on and compare these results with what would be obtained if reinforcement learning (RL) were used. The largest tests conducted here, which yielded very good results, involved barely 50 files. In contrast, RL would typically require thousands of examples or games to achieve comparable performance. This highlights the efficiency of LLMs in learning from a relatively small dataset, whereas RL approaches would need a significantly larger amount of data to train effectively. In future work, while our results indicate that RL is not a suitable alternative in this specific scenario, we aim to explore the potential benefits of integrating both approaches to enhance overall performance. However, it is important to note that the simulation environment used in this study is designed for test-time computation and is not conducive to on-policy learning methods.

\section{Acknowledgements}

This research was sponsored by the Department of the Air Force Artificial Intelligence Accelerator and was accomplished under Cooperative Agreement Number FA8750-19-2-1000. This work has also been supported by the MISTI-UPM program and by the Madrid Government (Comunidad de Madrid-Spain) under the Multiannual Agreement 2023-2026 with Universidad Politécnica de Madrid in the Line A, ``Emerging PhD researchers”. The views and conclusions contained in this document are those of the authors and should not be interpreted as representing the official policies, either expressed or implied, of the Department of the Air Force or the U.S. Government. The U.S. Government is authorized to reproduce and distribute reprints for Government purposes notwithstanding any copyright notation herein. Authors would like to thank Dr. Ross Allen and the rest of the team behind the development of the KSPDG challenge for the technical support and the creation of the competition.

\newpage
\appendix
\section{Detailed Weights \& Biases Projects}
\label{sec:detailed-wandb}
\begin{itemize}
    \item \textbf{LLaMA Experimental:}
    \begin{itemize}
        \item \textbf{Description:} This project contains the experimental runs and fine-tuning processes for the LLaMA models.
        \item \textbf{Link:} \href{https://wandb.ai/carrusk/LLaMA-Finetuning/workspace?nw=nwuserturnz7885}{LLaMA Experimental Project}
    \end{itemize}
    \item \textbf{LLaMA Final:}
    \begin{itemize}
        \item \textbf{Description:} This project includes the final versions of the LLaMA models with optimized parameters and performance metrics.
        \item \textbf{Link:} \href{https://wandb.ai/carrusk/huggingface/workspace?nw=nwuserturnz7885}{LLaMA Final Project}
    \end{itemize}
    \item \textbf{GPT Project:}
    \begin{itemize}
        \item \textbf{Description:} This project documents the experiments and results for the GPT models used in the KSPDG Challenge.
        \item \textbf{Link:} \href{https://wandb.ai/carrusk/KSPDG\%20Challenge\%20(paper)?nw=nwuserturnz7885}{GPT Project}
    \end{itemize}
\end{itemize}

\section{Detailed Hugging Face Models}
\label{sec:detailed-huggingface}
\begin{itemize}
    \item \textbf{LLaMA 35 files quantized:}
    \begin{itemize}
        \item \textbf{Link:} \href{https://huggingface.co/OhhTuRnz/train-KSPDG-Alpaca-LLaMA-Chat-8B-DoRA-16-8-adam_8bit_batch_2_gradient_2_LR_cosine_1e-4_8bit}{LLaMA 35 files quantized}
    \end{itemize}
    \item \textbf{LLaMA 10 files:}
    \begin{itemize}
        \item \textbf{Link:} \href{https://huggingface.co/OhhTuRnz/train-KSPDG-Alpaca-LLaMA-Chat-8B-DoRA-16-8-adamw_8bit_unquantized_10_files}{LLaMA 10 files}
    \end{itemize}
    \item \textbf{LLaMA 10 files sliding window:}
    \begin{itemize}
        \item \textbf{Link:} \href{https://huggingface.co/OhhTuRnz/train-KSPDG-Alpaca-LLaMA-Chat-8B-DoRA-16-8-adamw_8bit_unquantized_10_files_win_3}{LLaMA 10 files sliding window}
    \end{itemize}
    \item \textbf{LLaMA 10 files look ahead:}
    \begin{itemize}
        \item \textbf{Link:} \href{https://huggingface.co/OhhTuRnz/train-KSPDG-Alpaca-LLaMA-Chat-8B-DoRA-16-8-adamw_8bit_unquantized_10_files_la_3}{LLaMA 10 files look ahead}
    \end{itemize}
    \item \textbf{LLaMA 25 files:}
    \begin{itemize}
        \item \textbf{Link:} \href{https://huggingface.co/OhhTuRnz/train-KSPDG-Alpaca-LLaMA-Chat-8B-DoRA-16-8-adamw_8bit_unquantized_25_files}{LLaMA 25 files}
    \end{itemize}
    \item \textbf{LLaMA 50 files:}
    \begin{itemize}
        \item \textbf{Link:} \href{https://huggingface.co/OhhTuRnz/train-KSPDG-Alpaca-LLaMA-Chat-8B-DoRA-16-8-adamw_8bit_unquantized_50_files}{LLaMA 50 files}
    \end{itemize}
\end{itemize}

\definecolor{mygreen}{HTML}{10A982} 

\section{Prompting}
\label{sec:prompts}

\begin{figure}[h]
\centering
\fbox{
\begin{minipage}{.9\textwidth}
\small
\texttt{You operate as an autonomous agent controlling a pursuit spacecraft. Your goal is to apply throttles to capture the evader given the positions and velocities of the pursuer and evader in celestial body reference frame and the direction of pursuer's velocity relative to evader or prograde. Throttles must be given in your vessel reference frame wherein the x axis points to the right, the y axis points towards the target and the z axis points upwards. The maximum throttle is 1. Reason step-by-step.}
\end{minipage}
}
\caption{System prompt used both for the few-shot prompting and fine-tuning in LLaMA.}
\label{fig:system_prompt-llama-fine-tuning-and-inference}
\end{figure}

\begin{figure*}[h]
\centering
\fbox{
\begin{minipage}{.9\textwidth}
\small
\texttt{HUMAN: Given these observations ... ,what is the best throttle to capture evader? \newline ASSISTANT: The x coordinate of prograde is positive, indicating that pursuer is moving to the right. The y coordinate of prograde is negative, indicating that evader is approaching. The z coordinate of prograde is negative, indicating that pursuer is moving down. \textcolor{mygreen}{\textbf{To capture the evader we should counteract pursuer's motion, moving in the opposite direction in the x axis, towards the target in the y axis, and in the opposite direction in the z axis}}. This means we should apply throttles to move left, forward and up. Therefore we should call perform\_action({ft: forward, rt: left, dt: up}). \newline 
Now answer the following question: Given these observations ... , what is the best throttle to capture evader?. \textcolor{mygreen}{\textbf{Reason step-by-step.}}}
\end{minipage}
}
\caption{User prompt used both for the few-shot prompting and fine-tuning in LLaMA.}
\label{fig:user_prompt-llama-inference}
\end{figure*}

\section{Trajectories}
\label{sec:trajectories-appendix}

\begin{figure} [H]
\centering
\begin{subfigure}{.3\textwidth}
  \centering
  \includegraphics[scale=.35]{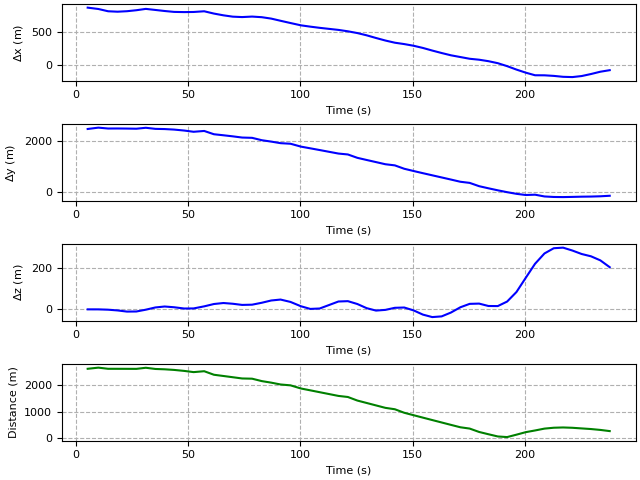}
  \caption{LLaMA 10 files}
  \label{fig:trajectory_10_files}
\end{subfigure}%
\hspace{5mm}
\begin{subfigure}{.3\textwidth}
  \centering
  \includegraphics[scale=.35]{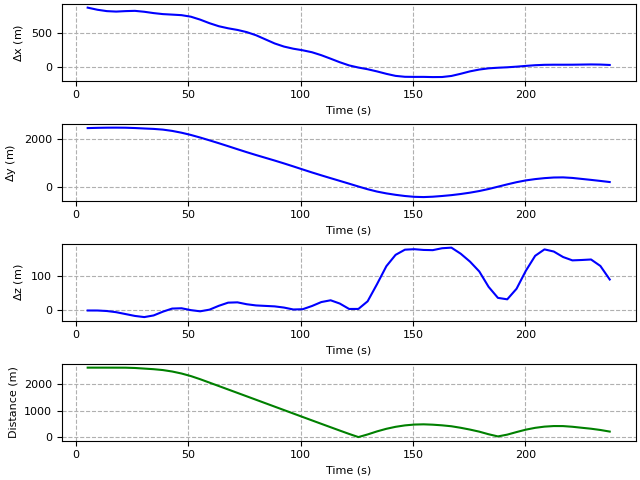}
  \caption{LLaMA 25 files}
  \label{fig:trajectory_25_files}
\end{subfigure}%
\hspace{5mm}
\begin{subfigure}{.3\textwidth}
  \centering
  \includegraphics[scale=.35]{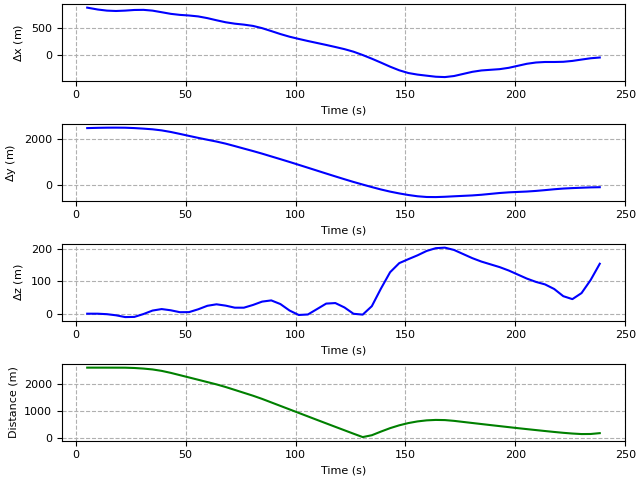}
  \caption{LLaMA 50 files}
  \label{fig:trajectory_50_files}
\end{subfigure}

\begin{subfigure}{.3\textwidth}
  \centering
  \includegraphics[scale=.35]{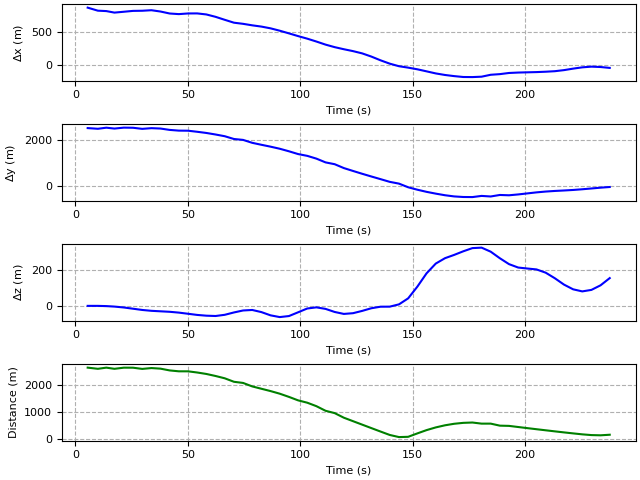}
  \caption{LLaMA 10 files win}
  \label{fig:trajectory_10_files_win}
\end{subfigure}%
\hspace{5mm}
\begin{subfigure}{.3\textwidth}
  \centering
  \includegraphics[scale=.35]{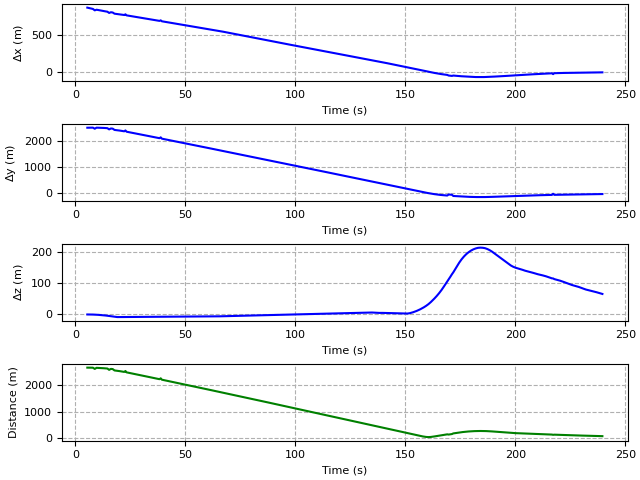}
  \caption{Navball bot}
  \label{fig:trajectory_navball}
\end{subfigure}

\caption[Behavior of the developed LLaMA models in Pursuer-Evader scenario E3]{Behavior of the developed LLaMA models in Pursuer-Evader scenario E3. The plots show the evolution of differences in position, velocity, as well as the relative distance and velocity metrics.}
\label{fig:trajectories_fine-tuned_llama}
\end{figure}

\begin{figure} [H]
\centering
\begin{subfigure}{.45\textwidth}
  \centering
  \includegraphics[scale=.35]{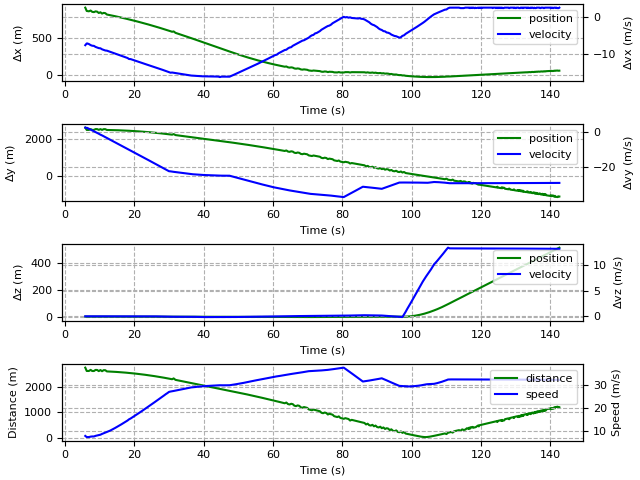}
  \caption{GPT agent based on observation augmentation and few-shot prompting (excluding CoT).}
  \label{fig:trajectory_few_shot}
\end{subfigure}%
\hspace{10mm}
\begin{subfigure}{.45\textwidth}
  \centering
  \includegraphics[scale=.35]{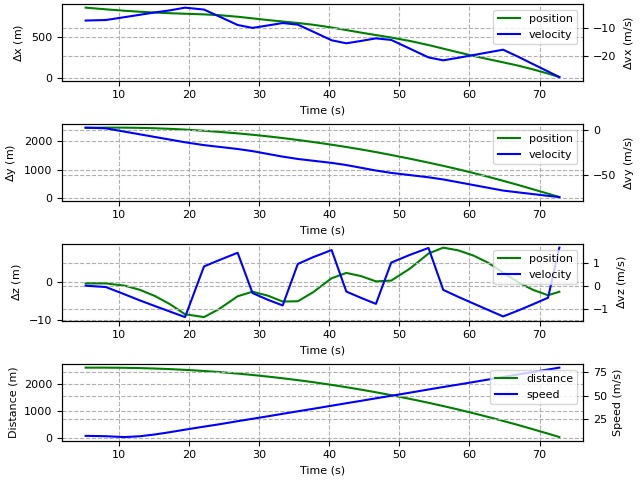}
  \caption{GPT agent based on fine-tuning.}
  \label{fig:trajectory_fine_tuning}
\end{subfigure}
\caption[Behavior of the developed GPT agents in Pursuer-Evader scenario E3]{Behavior of the developed GPT agents in Pursuer-Evader scenario E3. The plot shows the evolution of differences in the position, velocity, as well as the relative distance and velocity metrics.}
\label{fig:trajectories_grid_plot}
\end{figure}

\begin{figure} [H]
\centering
\includegraphics[scale=.30]{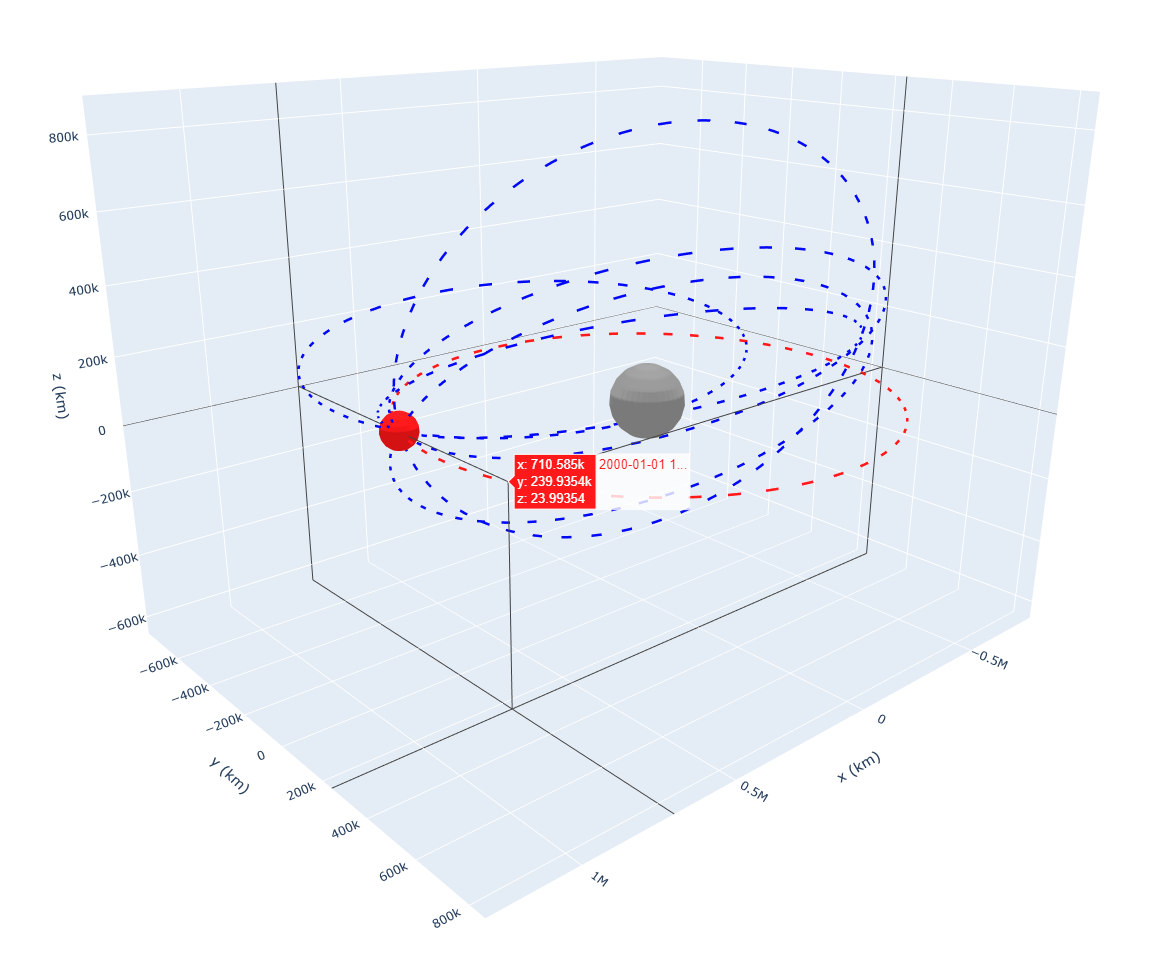}
\caption{An example of the generation of multiple orbits used to collect LLaMA training data}
\label{fig:multiple-orbits-generated}
\end{figure}

\section{Algorithms}
\label{sec:algorithms-appendix}

\begin{algorithm}[H]
\caption{LLM Agent Pseudocode}
\label{alg:LLMAgent}
\begin{algorithmic}[1]
\setlength{\parskip}{0.3em}
\State \textbf{Define class} LLMAgent(KSPDGBaseAgent):
\State \quad \textbf{Initialize} with scenario settings, prompt templates, and agent's configuration parameters like sliding window size
\State
\State \quad \textbf{Define} get\_action(observation)
\State \quad \quad Build new state with observations and augmented data (e.g. prograde)
\State \quad \quad Generate prompts (system / user) for new state
\State \quad \quad Add conversation history if sliding window is used
\State \quad \quad \textbf{Try}:
\State \quad \quad \quad Call LLM API
\State \quad \quad \quad \textbf{If} response includes function call \textbf{Then}
\State \quad \quad \quad \quad Extract action from function call
\State \quad \quad \quad \textbf{Else If} action can be extracted from response \textbf{Then}
\State \quad \quad \quad \quad Extract action from the response
\State \quad \quad \quad \textbf{Else}
\State \quad \quad \quad \quad Raise FailureException
\State \quad \quad \quad Append (state, action) pair to sliding window
\State \quad \quad \quad Log LLM interaction: state, action, input and output prompts
\State \quad \quad \quad Return action
\State \quad \quad \textbf{Except} (FailureException):
\State \quad \quad \quad Log the failure
\State \quad \quad \quad Wait for one second
\State \quad \quad \quad Return None or a default action
\State
\State \textbf{Main program:}
\State \quad Set scenario and environment
\State \quad Create instance of LLM Agent
\State \quad Create and run the environment's agent
\end{algorithmic}
\end{algorithm}

\begin{algorithm}[H]
\caption{Navball Agent Algorithm}
\begin{algorithmic}[1]
\setlength{\parskip}{0.3em}
\State \textbf{Define} get\_action(observation):
    \State Calculate prograde
    \quad \If {the deviation of the vessel's alignment from the prograde vector in the left-right or down-up directions exceeds ROTATION\_THRESHOLD}
        \State Apply rotational throttles to center the prograde marker on the navball
    \Else
        \State Do not apply rotation throttles
    \EndIf
    \State Calculate the braking distance if maximum backward throttle is applied, assuming uniform deceleration:
    \begin{equation}
    d = \frac{1}{2} \frac{v_0^2}{a}
    \end{equation}
    where $v_0$ is the initial relative speed and $a$ is VESSEL\_ACCELERATION.
    \If{the vessels are moving apart \textbf{or} the braking distance is less than the current distance}
        \State Apply forward throttle
    \Else
        \If{the approach speed exceeds APPROACH\_SPEED}
            \State Apply backward throttle
        \Else
            \State Do not apply thrust in the forward direction
        \EndIf
    \EndIf
    \State \textbf{return} action
\end{algorithmic}
\end{algorithm}

\begin{algorithm}[H]
\begin{algorithmic}[1]
\label{alg:orbit-generation-pseudocode}
\caption{Orbit Generation Pseudocode}
\setlength{\parskip}{0.3em}
\Repeat
    \State Randomly generate eccentricity, inclination, and initial distance within given constraints
    \State Estimate the radial distance as a fraction of the initial distance
    \State Calculate the semi-major axis considering that the radius is evader's orbit radius and correct it by adding the estimated radial distance. Where:
    \begin{equation}
    r = \frac{a (1 - e^2)}{1 + e \cos (\nu)}
    \label{eq:radius-vector}
    \end{equation}

    \begin{equation}
    v = \sqrt{\mu \left( \frac{2}{r} - \frac{1}{a} \right)}
    \end{equation}
    \State Set argument of periapsis and longitude of the ascending node to the evader's orbit argument of periapsis and true anomaly.
    \State Set true anomaly to zero
    \State Accept the generated orbit if it meets the distance constraints
\Until the desired number of orbits are generated
\end{algorithmic}
\end{algorithm}
\bibliographystyle{elsarticle-harv}
\bibliography{references} 

\end{document}